\newlength\footheight   

\documentclass{article}
\title{\vspace{-0.5in}The intersection of Finite State Automata and
Definite Clause Grammars}
\author{Gertjan van Noord\\
Vakgroep Alfa-informatica \& BCN\\
Rijksuniversiteit Groningen\\
{\tt vannoord@let.rug.nl}}

\usepackage{pictexwd}
\usepackage{aclap}

\begin{document}
\maketitle
\bibliographystyle{acl}
\vspace{-0.5in}
\begin{abstract}
  Bernard Lang defines parsing as the calculation of the
  intersection of a FSA (the input) and a CFG.  Viewing the input for
  parsing as a FSA rather than as a string combines well with some
  approaches in speech understanding systems, in which parsing takes
  a word lattice as input (rather than a word string).  Furthermore, certain
  techniques for robust parsing can be modelled as finite state
  transducers.

  In this paper we investigate how we can generalize this approach
  for unification grammars. In particular we will concentrate on how we
  might the calculation of the intersection of a FSA and a DCG. It is shown
  that existing parsing algorithms can be easily extended for FSA
  inputs. However, we also show that the termination properties change
  drastically: we show that it is undecidable whether the intersection
  of a FSA and a DCG is empty (even if the DCG is off-line parsable).

  Furthermore we discuss approaches to cope with the problem.
\end{abstract}

\section{Introduction}

In this paper we are concerned with the syntactic analysis phase of a
natural language understanding system. Ordinarily, the input of such a
system is a sequence of words. However, following Bernard Lang we
argue that it might be fruitful to take the input more generally
as a {\em finite state automaton (FSA)} to model cases in which we are
uncertain about the actual input. Parsing uncertain input might be
necessary in case of ill-formed textual input, or in case of speech input.

For example, if a natural language understanding system is interfaced
with a speech recognition component, chances are that this compenent
is uncertain about the actual string of words that has been uttered,
and thus produces a {\em word lattice} of the most promising
hypotheses, rather than a single sequence of words. FSA of course
generalizes such word lattices.

As another example, certain techniques to deal with ill-formed input
can be characterized as finite state transducers \cite{lang-atr}; the
composition of an input string with such a finite state transducer
results in a FSA that can then be input for syntactic parsing. Such an
approach allows for the treatment of missing, extraneous, interchanged or
misused words \cite{teitelbaum,saito,nederhof-bertsch}.

Such techniques might be of use both in the case of written and spoken
language input.   In the latter case another possible application concerns the
treatment of phenomena such as repairs \cite{carter-repairs}.

Note that we allow the input to be a full FSA (possibly including
cycles, etc.) since some of the above-mentioned techniques indeed
result in cycles. Whereas an ordinary word-graph always defines a
finite language, a FSA of course can easily define an infinite number
of sentences. Cycles might emerge to treat unknown sequences of words,
i.e. sentences with unknown parts of unknown lengths \cite{lang-coling88}.

As suggested by an ACL reviewer, one could also try to model haplology
phenomena (such as the 's in English sentences like `The chef at Joe's
hat', where `Joe's' is the name of a restaurant) using a finite state
transducer. In a  straightforward approach this would also lead to a
finite-state automaton with cycles.   \\

It can be shown that the computation of the intersection of a FSA
and a CFG requires only a minimal generalization of existing parsing
algorithms. We simply replace the usual string positions with the
names of the states in the FSA. It is also straightforward to
show that the complexity of this process is cubic in the number of
states of the FSA (in the case of ordinary parsing the number of
states equals $n+1$) \cite{lang74,billot-lang} (assuming the
right-hand-sides of grammar rules have at most two categories).

In this paper we investigate whether the same techniques can be
applied in case the grammar  is a constraint-based grammar rather
than a CFG. For specificity we will take the grammar to be a
{\em Definite Clause Grammar} (DCG) \cite{dcg}. A DCG is a simple
example of a family of constraint-based grammar formalisms that are widely
used in natural language analysis (and generation).
The main findings of this paper can be extended to other members
of that family of constraint-based grammar formalisms.

\section{The intersection of a CFG and a FSA}

The calculation of the intersection of a CFG and a FSA is very simple
\cite{bar-hillel}. The (context-free) grammar defining this
intersection is simply constructed by keeping track of the state names
in the non-terminal category symbols. For each rule $X_0 \rightarrow
X_1 \dots X_n$ there are rules $\langle X_0 q_0 q\rangle \rightarrow
\langle X_1 q_0 q_1\rangle \langle X_2 q_1 q_2\rangle \dots \langle
X_n q_{n-1} q\rangle $, for all $q_0 \dots q_n$. Furthermore for each
transition $\delta(q_i,\sigma)=q_k$ we have a rule $\langle\sigma q_i
q_k\rangle \rightarrow \sigma$. Thus the intersection of a FSA and a
CFG is a CFG that exactly derives all parse-trees. Such a grammar
might be called the parse-forest grammar.

Although this construction shows that the intersection of a FSA and a
CFG is itself a CFG, it is not of practical interest.  The reason is
that this construction typically yields an enormous amount of rules
that are `useless'. In fact the (possibly enormously large) parse
forest grammar might define an empty language (if the intersection was
empty). Luckily `ordinary' recognizers/parsers for CFG can be easily
generalized to construct this intersection yielding (in typical cases)
a much smaller grammar.  Checking whether the intersection is empty or
not is then usually very simple as well: only in the latter case will
the parser terminate succesfully.\\

To illustrate how a parser can be generalized to accept a FSA as input
we present a simple top-down parser.

A context-free grammar is represented as a definite-clause
specification as follows. We do not wish to define the sets of
terminal and non-terminal symbols explicitly, these can be understood
from the rules that are defined using the relation {\tt rule/2}, and
where symbols of the rhs are prefixed with `-' in the case of
terminals and `+' in the case of non-terminals. The relation {\tt
  top/1} defines the start symbol. The language $L'=a^nb^n$ is
defined as:

\small\begin{verbatim}
top(s).

rule(s,[-a,+s,-b]).  rule(s,[]).
\end{verbatim}\normalsize

In order to illustrate how ordinary parsers can be used to compute the
intersection of a FSA and a CFG consider first the definite-clause
specification of a top-down parser. This parser runs in polynomial
time if implemented using Earley deduction or XOLDT resolution
\cite{ds-warren}. It is assumed that the input string is represented
by the {\tt trans/3} predicate.

\small\begin{verbatim}
parse(P0,P) :-
    top(Cat), parse(+Cat,P0,P).

parse(-Cat,P0,P) :-
    trans(P0,Cat,P),
    side_effect(p(Cat,P0,P) --> Cat).
parse(+Cat,P0,P) :-
    rule(Cat,Ds),
    parse_ds(Ds,P0,P,His),
    side_effect(p(Cat,P0,P) --> His).

parse_ds([],P,P,[]).
parse_ds([H|T],P0,P,[p(H,P0,P1)|His]) :-
    parse(H,P0,P1),
    parse_ds(T,P1,P,His).
\end{verbatim}\normalsize

The predicate {\tt side\_effect} is used to construct the parse forest
grammar. The predicate always succeeds, and as a side-effect asserts
that its argument is a rule of the parse forest grammar. For the
sentence `a a b b' we obtain the parse forest grammar:
\small\begin{verbatim}
p(s,2,2) --> [].
p(s,1,3) -->
    [p(-a,1,2),p(+s,2,2),p(-b,2,3)].
p(s,0,4) -->
    [p(-a,0,1),p(+s,1,3),p(-b,3,4)].
p(a,1,2) --> a.
p(a,0,1) --> a.
p(b,2,3) --> b.
p(b,3,4) --> b.
\end{verbatim}\normalsize
The reader easily verifies that indeed this grammar generates (a
isomorphism of) the single parse tree of this example, assuming of
course that the start symbol for this parse-forest grammar is
{\tt p(s,0,4)}. In the parse-forest grammar, complex symbols are
non-terminals, atomic symbols are terminals. \\

\begin{figure*}[h,t,b,p]
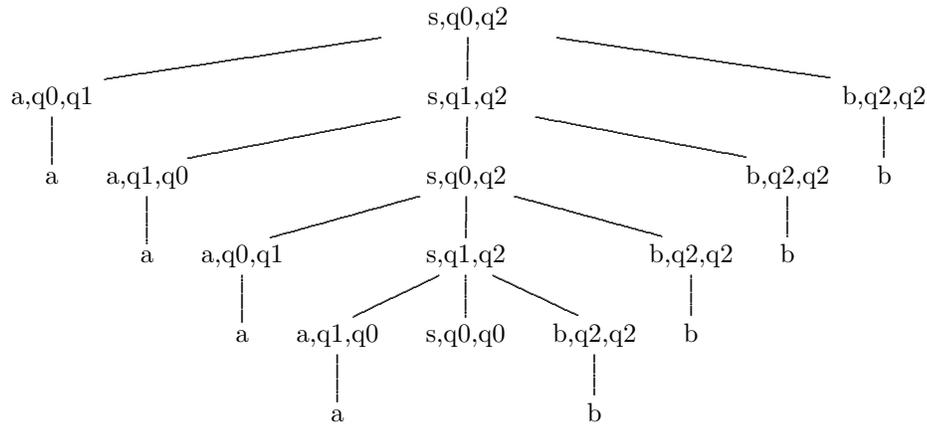

%
\begin{center}
\leavevmode
\unitlength1pt
\beginpicture
 \setplotarea x from -20 to 346.17, y from 0 to 186.89
\setlinear
\catcode`\@=11
 \put{\hbox{s,q0,q2}} [Bl] at 157.92 168.00
 \put{\hbox{}} [Bl] at 172.94 160.06
 \plot 139.9079 163.0556 35.6004 147.4444 / 
 \put{\hbox{a,q0,q1}} [Bl] at 0.00 138.00
 \put{\hbox{}} [Bl] at 15.56 130.06
 \plot 15.5556 133.0556 15.5556 115.3055 / 
 \put{\hbox{a}} [Bl] at 13.06 108.00
 \put{\hbox{}} [Bl] at 15.56 102.00
 \plot 172.8864 163.0556 172.7023 147.4444 / 
 \put{\hbox{s,q1,q2}} [Bl] at 157.64 138.00
 \put{\hbox{}} [Bl] at 172.67 130.06
 \plot 147.2393 133.0556 66.9561 117.4444 / 
 \put{\hbox{a,q1,q0}} [Bl] at 35.97 108.00
 \put{\hbox{}} [Bl] at 51.53 100.06
 \plot 51.5280 103.0556 51.5280 85.3055 / 
 \put{\hbox{a}} [Bl] at 49.03 78.00
 \put{\hbox{}} [Bl] at 51.53 72.00
 \plot 172.6087 133.0556 172.4245 117.4444 / 
 \put{\hbox{s,q0,q2}} [Bl] at 157.36 108.00
 \put{\hbox{}} [Bl] at 172.39 100.06
 \plot 154.5705 103.0556 98.3117 87.4444 / 
 \put{\hbox{a,q0,q1}} [Bl] at 71.94 78.00
 \put{\hbox{}} [Bl] at 87.50 70.06
 \plot 87.5004 73.0556 87.5004 55.3055 / 
 \put{\hbox{a}} [Bl] at 85.00 48.00
 \put{\hbox{}} [Bl] at 87.50 42.00
 \plot 172.3309 103.0556 172.1469 87.4444 / 
 \put{\hbox{s,q1,q2}} [Bl] at 157.08 78.00
 \put{\hbox{}} [Bl] at 172.11 70.06
 \plot 161.9020 73.0556 129.6672 57.4444 / 
 \put{\hbox{a,q1,q0}} [Bl] at 107.92 48.00
 \put{\hbox{}} [Bl] at 123.47 40.06
 \plot 123.4726 43.0556 123.4726 25.3055 / 
 \put{\hbox{a}} [Bl] at 120.97 18.00
 \put{\hbox{}} [Bl] at 123.47 12.00
 \plot 172.0824 73.0556 171.9903 57.4444 / 
 \put{\hbox{s,q0,q0}} [Bl] at 156.94 48.00
 \put{\hbox{}} [Bl] at 171.97 40.06
 \plot 182.5425 73.0556 214.4215 57.9444 / 
 \put{\hbox{b,q2,q2}} [Bl] at 204.92 48.00
 \put{\hbox{}} [Bl] at 220.75 40.06
 \plot 220.7504 43.0556 220.7504 27.9444 / 
 \put{\hbox{b}} [Bl] at 217.97 18.00
 \put{\hbox{}} [Bl] at 220.75 12.00
 \plot 190.5942 103.0556 246.2324 87.9444 / 
 \put{\hbox{b,q2,q2}} [Bl] at 241.44 78.00
 \put{\hbox{}} [Bl] at 257.28 70.06
 \plot 257.2782 73.0556 257.2782 57.9444 / 
 \put{\hbox{b}} [Bl] at 254.50 48.00
 \put{\hbox{}} [Bl] at 257.28 42.00
 \plot 198.6462 133.0556 278.0434 117.9444 / 
 \put{\hbox{b,q2,q2}} [Bl] at 277.97 108.00
 \put{\hbox{}} [Bl] at 293.81 100.06
 \plot 293.8060 103.0556 293.8060 87.9444 / 
 \put{\hbox{b}} [Bl] at 291.03 78.00
 \put{\hbox{}} [Bl] at 293.81 72.00
 \plot 206.6980 163.0556 309.8543 147.9444 / 
 \put{\hbox{b,q2,q2}} [Bl] at 314.50 138.00
 \put{\hbox{}} [Bl] at 330.33 130.06
 \plot 330.3339 133.0556 330.3339 117.9444 / 
 \put{\hbox{b}} [Bl] at 327.56 108.00
 \put{\hbox{}} [Bl] at 330.33 102.00
 \put{\hbox{}} [Bl] at 0.00 165.00
\endpicture
\end{center}
\caption{\label{pt}A parse-tree extracted from the parse forest grammar}
\end{figure*}

Next consider the definite clause specification of a FSA.
We define the transition relation using
the relation {\tt trans/3}. For start states, the relation start/1
should hold, and for final states the relation final/1 should hold.
Thus the following FSA, defining the regular language $L=(aa)^*b^+$
(i.e. an even number of a's followed by at least one b) is given as:\\

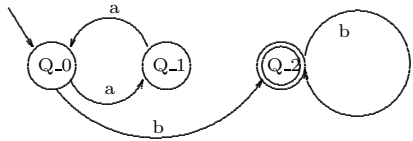
\begin{center}
\font\thinlinefont=cmr5
\begingroup\makeatletter\ifx\SetFigFont\undefined
\def\x#1#2#3#4#5#6#7\relax{\def\x{#1#2#3#4#5#6}}%
\expandafter\x\fmtname xxxxxx\relax \def\y{splain}%
\ifx\x\y   
\gdef\SetFigFont#1#2#3{%
  \ifnum #1<17\tiny\else \ifnum #1<20\small\else
  \ifnum #1<24\normalsize\else \ifnum #1<29\large\else
  \ifnum #1<34\Large\else \ifnum #1<41\LARGE\else
     \huge\fi\fi\fi\fi\fi\fi
  \csname #3\endcsname}%
\else
\gdef\SetFigFont#1#2#3{\begingroup
  \count@#1\relax \ifnum 25<\count@\count@25\fi
  \def\x{\endgroup\@setsize\SetFigFont{#2pt}}%
  \expandafter\x
    \csname \romannumeral\the\count@ pt\expandafter\endcsname
    \csname @\romannumeral\the\count@ pt\endcsname
  \csname #3\endcsname}%
\fi
\fi\endgroup
\mbox{\beginpicture
\setcoordinatesystem units < 0.400cm, 0.400cm>
\unitlength= 0.400cm
\linethickness=1pt
\setplotsymbol ({\makebox(0,0)[l]{\tencirc\symbol{'160}}})
\setshadesymbol ({\thinlinefont .})
\setlinear
%
%
\linethickness= 0.500pt
\setplotsymbol ({\thinlinefont .})
%
%
\plot  3.153 22.518  3.143 22.257  3.275 22.483 /
\circulararc 147.479 degrees from  5.683 22.257 center at  4.413 21.886
%
%
\linethickness= 0.500pt
\setplotsymbol ({\thinlinefont .})
%
%
\plot  9.409 20.898  9.493 21.145  9.302 20.967 /
\circulararc 108.924 degrees from  2.667 20.828 center at  5.967 23.425
%
%
\linethickness= 0.500pt
\setplotsymbol ({\thinlinefont .})
%
%
\plot 11.019 21.220 10.922 21.463 10.894 21.203 /
\circulararc 344.392 degrees from 10.922 21.463 center at 12.659 21.701
%
%
\linethickness= 0.500pt
\setplotsymbol ({\thinlinefont .})
\ellipticalarc axes ratio  0.794:0.794  360 degrees
        from  3.302 21.622 center at  2.508 21.622
%
%
\linethickness= 0.500pt
\setplotsymbol ({\thinlinefont .})
\ellipticalarc axes ratio  0.794:0.794  360 degrees
        from  7.112 21.622 center at  6.318 21.622
%
%
\linethickness= 0.500pt
\setplotsymbol ({\thinlinefont .})
\ellipticalarc axes ratio  0.794:0.794  360 degrees
        from 10.922 21.622 center at 10.128 21.622
%
%
\linethickness= 0.500pt
\setplotsymbol ({\thinlinefont .})
\ellipticalarc axes ratio  0.635:0.635  360 degrees
        from 10.763 21.622 center at 10.128 21.622
%
%
\linethickness= 0.500pt
\setplotsymbol ({\thinlinefont .})
%
%
\plot  5.486 20.887  5.524 21.145  5.369 20.935 /
\circulararc 135.086 degrees from  3.143 21.145 center at  4.334 21.638
%
%
\linethickness= 0.500pt
\setplotsymbol ({\thinlinefont .})
\plot  1.079 23.527  1.873 22.257 /
%
%
\plot  1.685 22.438  1.873 22.257  1.792 22.506 /
%
%
%
\put{\SetFigFont{7}{8.4}{rm}a} [lB] at  4.255 20.669
%
%
\put{\SetFigFont{7}{8.4}{rm}Q\_0} [lB] at  2.032 21.463
%
%
\put{\SetFigFont{7}{8.4}{rm}Q\_1} [lB] at  5.842 21.463
%
%
\put{\SetFigFont{7}{8.4}{rm}Q\_2} [lB] at  9.652 21.463
%
%
\put{\SetFigFont{7}{8.4}{rm}a} [lB] at  4.413 23.368
%
%
\put{\SetFigFont{7}{8.4}{rm}b} [lB] at 12.033 22.574
%
%
\put{\SetFigFont{7}{8.4}{rm}b} [lB] at  5.842 19.399
\linethickness=0pt
\putrectangle corners at  1.079 23.654 and 14.415 19.209
\endpicture}

\end{center}

\small\begin{verbatim}
start(q0).  final(q2).

trans(q0,a,q1).  trans(q1,a,q0).
trans(q0,b,q2).  trans(q2,b,q2).
\end{verbatim}\normalsize

Interestingly, nothing needs to be changed to use the same
parser for the computation of the intersection of a FSA and a CFG. If
our input `sentence' now is the definition of {\tt trans/3} as given
above, we obtain the following parse forest grammar (where the start
symbol is {\tt p(s,q0,q2)}):
\small\begin{verbatim}
p(s,q0,q0) --> [].
p(s,q1,q1) --> [].
p(s,q1,q2) -->
    [p(-a,q1,q0),p(+s,q0,q0),p(-b,q0,q2)].
p(s,q0,q2) -->
    [p(-a,q0,q1),p(+s,q1,q2),p(-b,q2,q2)].
p(s,q1,q2) -->
    [p(-a,q1,q0),p(+s,q0,q2),p(-b,q2,q2)].
p(a,q0,q1) --> a.
p(a,q1,q0) --> a.
p(b,q0,q2) --> b.
p(b,q2,q2) --> b.
\end{verbatim}\normalsize
Thus, even though we now use the same parser for an infinite set of
input sentences (represented by the FSA) the parser still is able to
come up with a parse forest grammar.  A possible derivation for this
grammar constructs the following (abbreviated) parse tree in figure~\ref{pt}.
Note that the construction of Bar Hillel would have yielded a grammar
with 88 rules.

\section{The intersection of a DCG and a FSA}
In this section we want to generalize the ideas described above for
CFG to DCG.

First note that the problem of calculating the intersection of a DCG
and a FSA can be solved trivially by a generalization of the
construction by \cite{bar-hillel}. However, if we use that method we
will end up (typically) with an enormously large forest grammar that
is not even guaranteed to contain solutions \footnote{In fact, the
  standard compilation of DCG into Prolog clauses does something
  similar using variables instead of actual state names. This also
  illustrates that this method is not very useful yet; all the work
  has still to be done.}. Therefore, we are
interested in methods that only generate a small subset of this; e.g.
if the intersection is empty we want an empty parse-forest grammar.

The straightforward approach is to generalize existing recognition
algorithms.
The same techniques that are used for calculating the intersection of
a FSA and a CFG can be applied in the case of DCGs. In order to
compute the intersection of a DCG and a FSA we assume that FSA are
represented as before. DCGs are represented using the same notation
we used for context-free grammars, but now of course the category
symbols can be first-order terms of arbitrary complexity (note that
without loss of generality we don't take into account DCGs
having external actions defined in curly braces).

\begin{figure*}[h,t,b,p]
\setlength{\unitlength}{.8pt}
\begin{picture}(400,100)(-100,0)

\put(0,0){\line(1,0){100}}
\put(0,5){\line(1,0){100}}
\put(0,0){\line(0,1){100}}
\put(0,50){\line(1,0){100}}
\put(100,100){\line(-1,0){100}}
\put(100,100){\line(0,-1){100}}
\put(50,25){\makebox(0,0){111}}
\put(50,75){\makebox(0,0){1}}
\put(10,90){\makebox(0,0){$A_1$}}
\put(10,40){\makebox(0,0){$B_1$}}

\put(125,0){\line(1,0){100}}
\put(125,5){\line(1,0){100}}
\put(125,0){\line(0,1){100}}
\put(125,50){\line(1,0){100}}
\put(225,100){\line(-1,0){100}}
\put(225,100){\line(0,-1){100}}
\put(175,25){\makebox(0,0){10}}
\put(175,75){\makebox(0,0){10111}}
\put(135,90){\makebox(0,0){$A_2$}}
\put(135,40){\makebox(0,0){$B_2$}}

\put(250,0){\line(1,0){100}}
\put(250,5){\line(1,0){100}}
\put(250,0){\line(0,1){100}}
\put(250,50){\line(1,0){100}}
\put(350,100){\line(-1,0){100}}
\put(350,100){\line(0,-1){100}}
\put(300,25){\makebox(0,0){0}}
\put(300,75){\makebox(0,0){10}}
\put(260,90){\makebox(0,0){$A_3$}}
\put(260,40){\makebox(0,0){$B_3$}}
\end{picture}

\caption{\label{pcp-ins} Instance of a PCP problem.}
\end{figure*}
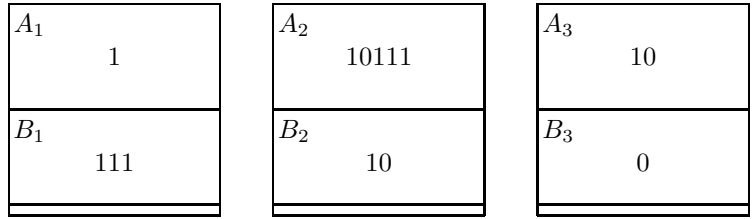

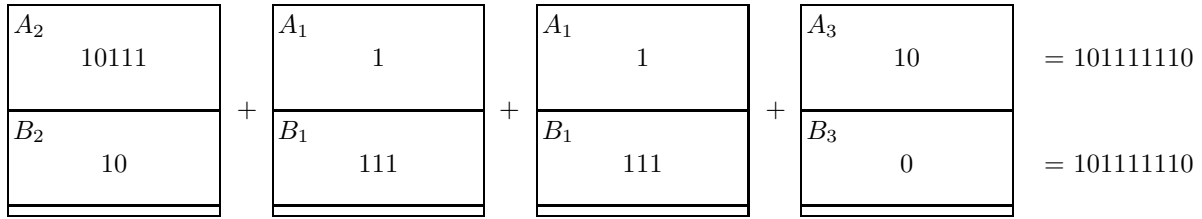
\begin{figure*}[h,t,b,p]
\setlength{\unitlength}{.8pt}
\begin{picture}(400,100)(0,0)
\put(0,0){\line(1,0){100}}
\put(0,5){\line(1,0){100}}
\put(0,0){\line(0,1){100}}
\put(0,50){\line(1,0){100}}
\put(100,100){\line(-1,0){100}}
\put(100,100){\line(0,-1){100}}
\put(50,25){\makebox(0,0){10}}
\put(50,75){\makebox(0,0){10111}}
\put(10,90){\makebox(0,0){$A_2$}}
\put(10,40){\makebox(0,0){$B_2$}}

\put(113,50){\makebox(0,0){+}}
\put(125,0){\line(1,0){100}}
\put(125,5){\line(1,0){100}}
\put(125,0){\line(0,1){100}}
\put(125,50){\line(1,0){100}}
\put(225,100){\line(-1,0){100}}
\put(225,100){\line(0,-1){100}}
\put(175,25){\makebox(0,0){111}}
\put(175,75){\makebox(0,0){1}}
\put(135,90){\makebox(0,0){$A_1$}}
\put(135,40){\makebox(0,0){$B_1$}}

\put(237,50){\makebox(0,0){+}}
\put(250,0){\line(1,0){100}}
\put(250,5){\line(1,0){100}}
\put(250,0){\line(0,1){100}}
\put(250,50){\line(1,0){100}}
\put(350,100){\line(-1,0){100}}
\put(350,100){\line(0,-1){100}}
\put(300,25){\makebox(0,0){111}}
\put(300,75){\makebox(0,0){1}}
\put(260,90){\makebox(0,0){$A_1$}}
\put(260,40){\makebox(0,0){$B_1$}}

\put(363,50){\makebox(0,0){+}}
\put(375,0){\line(1,0){100}}
\put(375,5){\line(1,0){100}}
\put(375,0){\line(0,1){100}}
\put(375,50){\line(1,0){100}}
\put(475,100){\line(-1,0){100}}
\put(475,100){\line(0,-1){100}}
\put(425,25){\makebox(0,0){0}}
\put(425,75){\makebox(0,0){10}}
\put(385,90){\makebox(0,0){$A_3$}}
\put(385,40){\makebox(0,0){$B_3$}}
\put(525,25){\makebox(0,0){= 101111110}}
\put(525,75){\makebox(0,0){= 101111110}}
\end{picture}
\caption{\label{pcp-xx} Illustration of a solution for the PCP problem
of figure~\ref{pcp-ins}.}
\end{figure*}

But if we use existing techniques for parsing DCGs, then we are also
confronted with an undecidability problem: the recognition problem for
DCGs is undecidable \cite{parsing-as-deduction}. A fortiori the
problem of deciding whether the intersection of a FSA and a DCG is
empty or not is undecidable.

This undecidability result is usually circumvented by considering
subsets of DCGs which can be recognized effectively. For example, we
can restrict the attention to DCGs of which the context-free skeleton
does not contain cycles. Recognition for such `off-line parsable'
grammars is decidable \cite{parsing-as-deduction}.

Most existing constraint-based parsing algorithms will terminate for
grammars that exhibit the property that for each string there is only
a finite number of possible derivations. Note that off-line
parsability is one possible way of ensuring that this is the case.

This observation is not very helpful in establishing insights
concerning interesting subclasses of DCGs for which termination can
be guaranteed (in the case of FSA input).
The reason is that there are now two sources of
recursion: in the DCG and in the FSA (cycles). As we saw earlier:
even for CFG it holds that there can be an infinite number of analyses
for a given FSA (but in the CFG this of course does not imply
undecidability).  \\

\begin{figure*}[h,t,b,p]

\small\begin{verbatim}
          trans(q0,x,q0).   start(q0).   final(q0).   % FSA

          top(s).                                     % start symbol DCG

          rule(s,[-r(X,[],X,[])]).                    % require A's and B's
%%match

          rule(r(A0,A,B0,B),[-r(A0,A1,B0,B1),         % combine two sequences
%%of
                             -r(A1,A,B1,B)]).         % blocks

          rule(r([1|A],        A,[1,1,1|B],B),[+x]).  % block A1/B1
          rule(r([1,0,1,1,1|A],A,[1,0|B],  B),[+x]).  % block A2/B2
          rule(r([1,0|A],      A,[0|B],    B),[+x]).  % block A3/B3
\end{verbatim}\normalsize

\caption{\label{x23}The encoding for the PCP problem of
  figure~\ref{pcp-ins}. }
\end{figure*}

\subsection{Intersection of FSA and off-line parsable DCG is undecidable}
I now show that the question whether the intersection of a FSA and an
off-line parsable DCG is empty is undecidable.  A yes-no problem is
{\em undecidable} (cf.  \cite[pp.178-179]{hopcroft-ullman}) if there
is no algorithm that takes as its input an {\em instance} of the
problem and determines whether the answer to that instance is `yes' or
`no'.  An instance of a problem consists of a particular choice of the
{\em parameters} of that problem.

I use Post's Correspondence Problem (PCP) as a well-known undecidable
problem. I show that if the above mentioned intersection problem were
decidable, then we could solve the PCP too.
The following definition and example of a PCP are taken from
\cite{hopcroft-ullman}[chapter 8.5].

An instance of PCP consists of two lists, $A = v_1 \dots v_k$ and $B =
w_1 \dots w_k$ of strings over some alphabet $\Sigma$. This instance
has {\em a solution} if there is any sequence of integers $i_1 \dots
i_m$, with $m \geq 1$, such that
\[
v_{i_1}, v_{i_2}, \dots, v_{i_m} = w_{i_1}, w_{i_2}, \dots, w_{i_m}.
\]\noindent
The sequence $i_1, \dots, i_m$ is a solution to this instance of PCP.
As an example, assume that $\Sigma = \{0,1\}$. Furthermore, let
$A = \langle 1,10111,10 \rangle$ and $B = \langle 111, 10, 0 \rangle$. A
solution to this instance of PCP is the sequence
2,1,1,3 (obtaining the sequence 101111110). For an illustration, cf.
figure~\ref{pcp-xx}.

Clearly there are PCP's that do not have a solution. Assume again that
$\Sigma = \{0,1\}$. Furthermore let $A = \langle 1 \rangle$ and $B =
\langle 0 \rangle$. Clearly this PCP does not have a solution.  In
general, however, the problem whether some PCP has a solution or not
is not decidable.  This result is proved by
\cite{hopcroft-ullman} by showing that the halting problem for
Turing Machines can be encoded as an instance of Post's Correspondence
Problem.

First I give a simple algorithm to encode any instance of a PCP as a
pair, consisting of a FSA and an off-line parsable DCG,
in such a way that the question whether there is a
solution to this PCP is equivalent to the question whether the
intersection of this FSA and DCG is empty.

\paragraph{Encoding of PCP.}
\begin{enumerate}
\item For each $1 \leq i \leq k$ ($k$ the length of lists $A$ and $B$)
define a DCG rule (the $i-th$ member of $A$ is $a_1 \dots
a_m$, and the $i-$th member of $B$ is $b_1 \dots b_n$):
$r([a_1\dots a_m|A],A,[b_1\dots b_n|B],B) \rightarrow [x].$
\item Furthermore, there is a rule
$ r(A_0,A,B_0,B) \rightarrow r(A_0,A_1,B_0,B_1), r(A_1,A,B_1,B).$
\item Furthermore, there is a rule
$ s \rightarrow r(X,[~],X,[~]).$ Also, $s$ is the start category of the
DCG.
\item Finally, the FSA consists of a single state $q$ which is both
  the start state and the final state, and a single transition
$\delta(q,x)=q$. This FSA generates $x^*$.
\end{enumerate}
Observe that the DCG is off-line parsable.

The underlying idea of the algorithm is really very simple. For each pair
of strings from the lists A and B there will be one lexical entry
(deriving the terminal $x$) where these
strings are represented by a difference-list encoding. Furthermore there
is a general combination rule that simply concatenates A-strings and
concatenates B-strings. Finally the rule for $s$ states that in order
to construct a succesful top category the A and B lists must match.

The resulting DCG, FSA pair for the example PCP is given in figure~\ref{x23}:

\paragraph{Proposition}
The question whether the intersection of a FSA and an off-line
parsable DCG is empty is undecidable.

\paragraph{Proof.} Suppose the problem {\em was} decidable. In that
case there would exist an algorithm for solving the problem. This
algorithm could then be used  to solve the PCP,
because a PCP $\pi$ has a solution
if and only if its encoding given above as a FSA and an off-line
parsable DCG is not empty.
The PCP problem however is known to be undecidable. Hence the
intersection question is undecidable too.

\subsection{What to do?}

The following approaches towards the undecidability problem can be
taken:

\begin{itemize}
\item limit the power of the FSA
\item limit the power of the DCG
\item compromise completeness
\item compromise soundness
\end{itemize}

These approaches are discussed now in turn.

\paragraph{Limit the FSA} Rather than assuming the input for parsing
is a FSA in its full generality, we might assume that the input is
an ordinary word graph (a FSA without cycles).

Thus the techniques for robust processing that give rise to such
cycles cannot be used. One example is the processing of an unknown
sequence of words, e.g. in case there is noise in the input and it is
not clear how many words have been uttered during this noise.
It is not clear to me right now what we loose (in practical terms) if
we give up such cycles. \\

Note that it is easy to verify that the question whether the
intersection of a word-graph and an off-line parsable DCG is empty or
not is decidable since it reduces to checking whether the DCG derives
one of a finite number of strings.

\paragraph{Limit the DCG}
Another approach is to limit the size of the categories that are being
employed. This is the GPSG and F-TAG approach. In that case we are not
longer dealing with DCGs but rather with CFGs (which have been shown to
be insufficient in general for the description of natural languages).

\paragraph{Compromise completeness} Completeness in this context
means: the parse forest grammar contains all possible parses. It is
possible to compromise here, in such a way that the parser is
guaranteed to terminate, but sometimes misses a few parse-trees.

For example, if we assume that each edge in the FSA is associated with
a probability it is possible to define a thres\-hold such that each
partial result that is derived has a probability higher than the
thres\-hold. Thus, it is still possible to have cycles in the FSA, but
anytime the cycle is `used' the probability decreases and if too many
cycles are encountered the thres\-hold will cut off that derivation.

Of course this implies that sometimes the intersection is considered
empty by this procedure whereas in fact the intersection is not.  For
any thres\-hold it is the case that the intersection problem of off-line
parsable DCGs and FSA is decidable.

\paragraph{Compromise soundness} Soundness in this context should be
understood as the property that all parse trees in the parse forest
grammar are valid parse trees. A possible way to ensure termination
is to remove all constraints from the DCG and parse
according to this context-free skeleton. The resulting parse-forest
grammar  will be too general most of the times.

A practical variation can be conceived as follows. From the DCG we
take its context-free skeleton. This skeleton is obtained by removing
the constraints from each of the grammar rules. Then we compute the
intersection of the skeleton with the input FSA. This results in a
parse forest grammar. Finally, we add the corresponding constraints
from the DCG to the grammar rules of the parse forest grammar.

This has the advantage that the result is still sound and complete,
although the size of the parse forest grammar is not optimal (as a
consequence it is not guaranteed that the parse forest grammar
contains a parse tree). Of course it is possible to experiment with
different ways of taking the context-free skeleton (including as much
information as possible / useful).

\section*{Acknowledgments}
I would like to thank Gosse Bouma, Mark-Jan Nederhof and John Nerbonne for
comments on this paper. Furthermore the paper benefitted from
remarks made by the anonymous ACL reviewers.

\end{document}